\documentclass[journal]{IEEEtran}
\usepackage{xcolor,soul,framed}
\colorlet{shadecolor}{yellow}
\usepackage{graphicx}
\graphicspath{{figures/}}
\usepackage{amsmath,amssymb,amsfonts}
\usepackage{array}
\usepackage{mdwmath}
\usepackage{mdwtab}
\usepackage{eqparbox}
\usepackage{url}
\usepackage{bm}
\usepackage{enumerate}
\usepackage{float}
\usepackage{nomencl}
\usepackage{etoolbox}
\usepackage{mathrsfs}
\usepackage{booktabs}
\usepackage{tabularx}
\usepackage{fancybox}
\usepackage{subfig}
\usepackage[ruled,linesnumbered,vlined]{algorithm2e}
\usepackage{bbm}
\usepackage{multirow}
\usepackage{makecell}
\usepackage{hyperref} 

\begin{document}

\title{Privacy-preserving Decision-focused Learning for Multi-energy Systems}

\author{Yangze Zhou, Ruiyang Yao, Dalin Qin, Yixiong Jia and Yi Wang
        
}

\markboth{Submitted to IEEE Trans. Smart Grid}%
{Shell \MakeLowercase{\textit{et al.}}: Bare Demo of IEEEtran.cls for IEEE Journals}
\maketitle

\begin{abstract}
Decision-making for multi-energy system (MES) dispatch depends on accurate load forecasting.
Traditionally, load forecasting and decision-making for MES are implemented separately. Forecasting models are typically trained to minimize forecasting errors, overlooking their impact on downstream decision-making. To address this, decision-focused learning (DFL) has been studied to minimize decision-making costs instead. However, practical adoption of DFL in MES faces significant challenges: the process requires sharing sensitive load data and model parameters across multiple sectors, raising serious privacy issues. To this end, we propose a privacy-preserving DFL framework tailored for MES. Our approach introduces information masking to safeguard private data while enabling recovery of decision variables and gradients required for model training. To further enhance security for DFL, we design a safety protocol combining matrix decomposition and homomorphic encryption, effectively preventing collusion and unauthorized data access. 
Additionally, we developed a privacy-preserving load pattern recognition algorithm, enabling the training of specialized DFL models for heterogeneous load patterns. Theoretical analysis and comprehensive case studies, including real-world MES data, demonstrate that our framework not only protects privacy but also consistently achieves lower average daily dispatch costs compared to existing methods.
\end{abstract}

\IEEEpeerreviewmaketitle




\section{Introduction}
\IEEEPARstart{D}riven by concerns over global warming and the exhaustion of fossil energy resources, increasing attention is being paid to the decarbonization of energy systems \cite{lu2018coordinated}. Multi-energy systems (MES) integrate electricity, heat, cooling, gas, and other energy carriers within a single framework, covering production, transportation, storage, and consumption \cite{li2022optimal}.
By utilizing the complementary features of different energy sectors, MES improve energy efficiency and resource management \cite{guelpa2019towards}. Thus, MES are considered a promising pathway for future low-carbon energy systems \cite{mancarella2014mes}. The energy management system ensures the safe and efficient operation of MES.
It relies on a multi-stage dispatch process, such as day-ahead and intra-day dispatch \cite{sun2018integrated}. Among these,
day-ahead dispatch sets the foundation by planning daily schedules and operating plans. Later stages, such as intra-day dispatch, adjust these plans as forecasts and real conditions change. Consequently, day-ahead dispatch is a key focus in MES research \cite{sun2018integrated}.
To address the diverse needs and complexities of MES, various day-ahead dispatch strategies have been proposed. For example, \cite{zhang2021day} developed a day-ahead dispatch strategy for MES that considered power-to-gas and dynamic pipeline networks. \cite{yong2021day} addressed the non-linear energy efficiency of coupling equipment under varying operating conditions. In addition, \cite{zhang2023day} proposed an efficient day-ahead dispatch approach by leveraging both conventional units and energy storage to provide multi-timescale regulation services.

The effectiveness of these MES day-ahead dispatch strategies critically depends on accurate multi-energy load forecasting (MELF), which provides essential input for scheduling decisions \cite{zhu2022review}. There are two primary MELF methods in the current research \cite{zhou2023can}. The first one is single-load forecasting, which is independently applied to each energy sector, such as electricity, heat, and cooling. The second approach accounts for the interdependencies among different energy loads; for instance, higher heating demand in winter often coincides with lower cooling demand. To capture these complex correlations, advanced techniques such as feature selection \cite{zhou2023can} and model structure design \cite{wu2024multi} 
have been developed, significantly improving MELF accuracy.
However, most existing MELF approaches train forecasting models using loss functions such as mean absolute error (MAE) or mean squared error (MSE), without considering the operational impact of forecasting errors on MES dispatch. In practice, positive and negative forecasting errors may have asymmetric effects on dispatch costs and system reliability \cite{li2016toward}. 

To address this gap, recent research has proposed decision-focused learning (DFL) frameworks, which integrate the dispatch decision process into the training of forecasting models. By optimizing forecasting outputs for downstream decision quality, DFL better aligns forecasting with MES decision-making, enabling MES to achieve more cost-effective and reliable dispatch.
For instance, with the optimal piecewise linear approximation method, \cite{zhang2022cost} derived a differentiable cost-oriented loss function to reflect the cost relative to forecasting errors.
\cite{han2021task} proposed a task-based day-ahead load forecasting method, which combined stochastic economic dispatch with probabilistic forecasting.
\cite{elmachtoub2022smart} proposed a ``smart predict then optimize" framework, and \cite{sang2022electricity} introduced it to price forecasting for energy storage systems to obtain better economic benefits. \cite{yi2025perturbed} proposed DFL
based on a perturbed method for modeling the energy storage bidding strategy.
\cite{zhao2021cost} formulated the prediction intervals of renewable energy as bi-level programming, enabling the joint improvement of forecasting quality and decision performance. 
\cite{zhang2024toward} proposed DFL renewable energy forecasting with an iterative learning approach.
Existing DFL research has primarily focused on power systems and energy storage systems, while studies on DFL in MES remain very limited. \cite{zhou2024load} integrates MES dispatch with forecasting model training to reduce system dispatch costs.
When constructing DFL models for MES decision-making, there are two potential challenges.
\begin{enumerate}
    \item \textbf{Privacy-leakage issues: }Implementing DFL in MES dispatch is particularly challenging due to the strong coupling among various sectors. Decision-making in MES typically requires sharing original load data and model parameters from different sectors for central computation. However, parameters related to diverse equipment and loads, spanning electricity, gas, and heating/cooling systems, are typically managed by separate system operators \cite{wang2023data}, which raises significant privacy concerns.
    \item \textbf{Load profiles heterogeneity: } Due to the inter-coupling of different energy forms in MES, multi-energy loads exhibit strong correlations and seasonality, resulting in diverse load patterns across various periods. Most existing studies overlook the heterogeneity of load profiles and train a single DFL model for all load patterns, which may lead to suboptimal performance. To consider the strong coupling in MES, existing algorithms demand load data from various sectors for centralized processing, which leads to privacy concerns.
    How to recognize diverse load patterns and develop an adaptive DFL framework that effectively handles load heterogeneity within a privacy-preserving context remains an open challenge.
\end{enumerate}
To bridge these gaps, we propose a privacy-preserving DFL framework that explicitly considers the heterogeneity of load profiles. The key contributions of this paper are summarized as follows.
\begin{enumerate}
    \item Studied a topic that is rarely noticed in DFL, potential privacy leakage arising from information exchange among various sectors in MES. This work proposes a privacy-preserving DFL framework that integrates forecasting and decision-masking for MES. Specifically, our approach introduces an information masking (IM) mechanism that reformulates the original optimization problem into a masked version, ensuring sector-specific parameter protection. Each sector can locally recover gradients of cost overs forecasts for DFL, without exposing sensitive information. The theoretical analysis confirms the effectiveness of the proposed approach.
    \item Design a safety protocol for the privacy-preserving DFL framework to prevent unauthorized queries and collusion between curious clients and the server. This protocol utilizes matrix blocking, where sectors generate the masking matrices independently. Furthermore, homomorphic encryption (HE) is incorporated to ensure the privacy of gradient information. This guarantees that intercepted communication cannot be decrypted by adversaries.
    \item An adaptive DFL method is proposed within the privacy-preserving framework to address load pattern heterogeneity. This approach introduces a privacy-preserving load pattern recognition algorithm (LPR), leveraging Euclidean space orthogonal transformations to enable IM-based K-means clustering. Experimental results indicate that this adaptive DFL, which targets individual load patterns, can decrease decision-masking costs compared to training a single DFL model for all load patterns.
\end{enumerate}

The rest of the paper is structured as follows. Section \ref{PS} formulates the problem statement. Section \ref{preliminaries} introduces the fundamental concepts and techniques relevant to this study. Section \ref{Methodology} details the proposed privacy-preserving DFL framework for MES. Section \ref{CS} analyzes and visualizes the experiment results. Finally, Section \ref{conclusion} concludes the paper.

\section{Problem Statement}
\label{PS}
Assume we are studying the following scenario:
there are $n$ sectors in the studied MES and each sector $i\in [1,n]$ has a forecasting model $F_i$ with parameters $w_i$. 
Traditionally, the forecasting model $F_i$ is trained to minimize forecasting errors using metrics that measure forecasting accuracy, such as MAE and MSE. 
As for decision-making in MES, each sector $i$ will report its forecasts $\hat{l}_i=F_i(X_i,w_i)$ with a length of $\tau$, to the server. 
Then, the server computes the decision variables $z$
according to received all forecasts (denoted as $\hat{l}$).

In this work, we formulate the MES decision-making problem as a mixed-integer linear or quadratic programming (MILP/QP) problem, which is a mainstream approach for such scheduling tasks \cite{wang2018automatic}.
The decision variables $z$ can be divided into continuous variables $z^{\text{T}}_C \in \mathbb{R}^{m_C}$ and integer variables $z^{\text{T}}_I \in \mathbb{R}^{m_I}$, where $m_C$ and $m_I$ are the number of these variables \cite{li2018optimal}.
For convenience, we still denote the decision variables that contain slack variables by $z_C$ and $z_I$ as \eqref{original_prob}. It is noted that although complex nonlinear constraints or objective functions exist, such as thermal dynamic characteristics or nonlinear generation outputs, these can typically be reformulated as MILP/QP problems \cite{mallegol2023handling}. This consideration is particularly important because directly solving nonlinear, nonconvex problems is often computationally intractable and may not guarantee solution accuracy within reasonable time frames. Consequently, approximating nonlinear functions with piecewise linear representations has been widely adopted to reformulate such problems as MILP/QP. 
For instance, \cite{yong2021day} handled nonlinear energy efficiency by applying piecewise linear approximations, thereby converting a non-convex problem into an MILP formulation. Similarly, \cite{shao2016milp} transformed nonconvex natural gas transmission constraints into linear constraints and provided theoretical guarantees on the approximation accuracy. 
\begin{subequations}\label{original_prob}
\begin{alignat}{4}
 \min_{z_C,z_I} \quad &\frac{1}{2}z_C^{\text{T}}Hz_C+d_C^{\text{T}}z_C+d_I^{\text{T}}z_I\\ 
  s.t. \quad & Iz_C\geq 0 \label{1b}\\
      & Mz_C+Nz_I=b \label{1c}\\ 
&M_{\text{EB}}z_C+N_{\text{EB}}z_I=\hat{l} \label{1d}
\end{alignat}
\label{straincomponent}
\end{subequations}
where $H\in \mathbb{R}^{m_C\times m_C}$, $d_C\in \mathbb{R}^{m_C}$, $d_I\in \mathbb{R}^{m_I}$ are the parameters related to costs of various device,  \eqref{1c} denotes device operation constraints, such as maximum and minimum output, and $M\in \mathbb{R}^{n_b\times m_C}$, $N\in \mathbb{R}^{n_b\times m_I}$, $b\in \mathbb{R}^{n_b}$ are the related parameters where $n_b$ is the number of device operation constraints,
\eqref{1d} is energy balance constraints and $M_{\text{EB}}\in \mathbb{R}^{n_\tau\times m_C}$, $N_{\text{EB}}\in \mathbb{R}^{n_\tau\times m_I}$ are the related parameters.

However, this approach has one limitation: forecasting and decision-making for MES are treated as separate processes. 
It is important to note that achieving lower forecasting errors does not necessarily reduce operating costs \cite{zhang2022cost}.
To this end, a possible solution is to integrate forecasting and decision-making via a DFL approach. In this approach, the decision variables $z_{C},z_{I}$ and forecasting model parameters $w_i,i\in[1,n]$ are simultaneously solved to minimize the dispatch costs. However, there are two drawbacks of the existing DFL:
\begin{enumerate}
    \item Implying existing DFL in MES typically requires sharing original load data ($\hat{l}$) and model parameters (e.g., $M$ and $N$) from different sectors for a central computation. However, these data are obtained by different data owners in MES. Hence, the first task of this work is to address the question:
    \textit{How can DFL be implemented in MES without direct sharing of raw load data or model parameters?} 
    \item Multi-energy loads demonstrate significant seasonality, showcasing diverse load patterns that fluctuate throughout different seasons. However, current DFL typically trains a single model for load profiles from different patterns, ignoring the heterogeneity of load profiles' impact on the training of DFL models. Hence, the second task of this work is to answer: 
    \textit{How can load patterns be identified and specialized DFL models be trained to address load heterogeneity under privacy constraints?}
\end{enumerate}



\begin{figure}[t]
\centering
\includegraphics[width=0.5\textwidth]{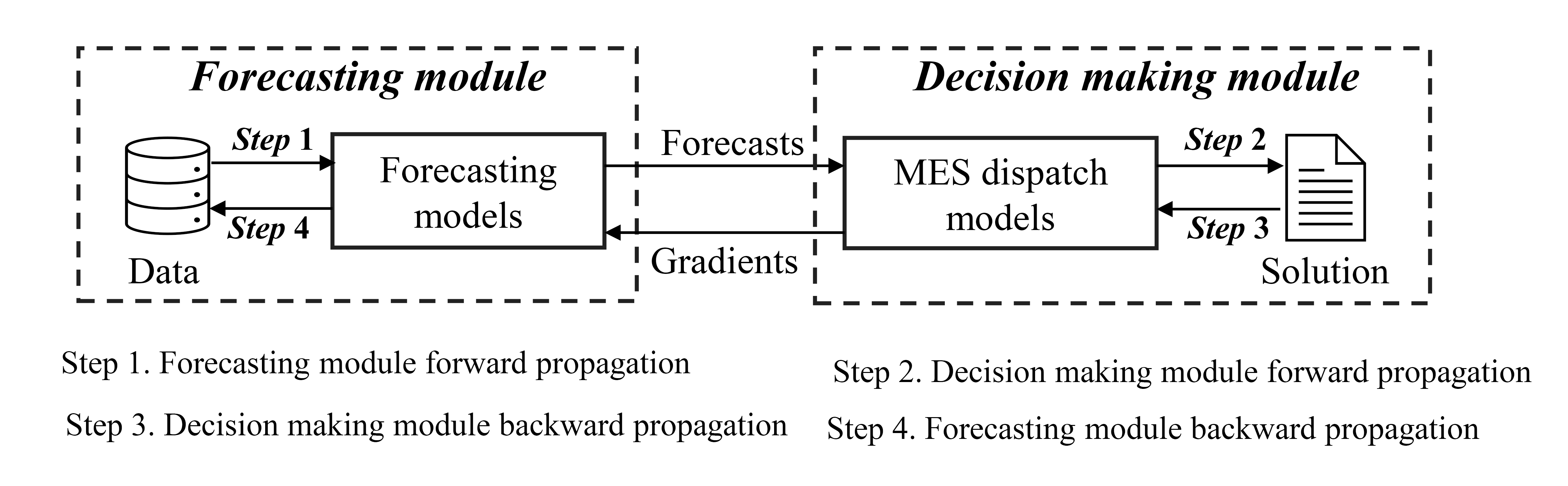}
\caption{Illustration of DFL for MES}
\label{framework}
\end{figure}
\section{Preliminaries}\label{preliminaries}
\subsection{Optimization Differentiable Neural Network}
As illustrated in Fig. \ref{framework}, an intuitive approach to conducting DFL involves forward and backward propagation, which is commonly used for traditional neural network training. 
In standard DFL, Steps 1, 2, and 4 can be easily implemented, while the main challenge lies in Step 3. This section introduced this step as preliminaries of our work.

The backward propagation aims to guide the training of the forecasting module with the gradient of optimal dispatch cost $O^{*}$ over forecasts $\hat{l}$ (desired gradient). 
For QP/LP problems with continuous decision variables, OptNet \cite{amos2017optnet} provides an efficient way to compute this gradient. Such problems can be formulated as:
\begin{subequations}
\begin{alignat}{3}
\min_{z_C}  \quad &\mathcal{O}(z_C,\hat{l})\\ 
      s.t.  \quad &\mathcal{F}_\text{eq}(z_C,\hat{l})=0,~
 &\mathcal{F}_\text{iq}(z_C,\hat{l})\leq0
\end{alignat}
\label{straincomponent}
\end{subequations}
The Lagrangian function of the optimization problem is
\begin{equation}
    L(z_C,\hat{l},\lambda,\mu)=\mathcal{O}(z_C,\hat{l})+\lambda \mathcal{F}_\text{iq}(z_C,\hat{l})+\mu \mathcal{F}_\text{eq}(z_C,\hat{l})
\end{equation}
where $\lambda$ and $\mu$ are dual variables. For such a convex and continuous problem, KKT conditions are sufficient and necessary conditions for optimality. With the stationarity condition,
primal feasibility, and the complementary slackness condition for the inequality constraints in the KKT condition, we can obtain an implicit function $\mathcal{K}$, denoting $[{z}^{*}_C,\lambda,\mu]$ as $\widetilde{z}$:
\begin{equation}
\begin{aligned}
        \mathcal{K}(\widetilde{z},\hat{l})=\left[
               \begin{array}{c}
\nabla_{{z}^{*}_C} \mathcal{L}({z}^{*}_C,\hat{l},\lambda,\mu)\\
                \lambda\mathcal{F}_\text{iq}({z}^{*}_C,\hat{l})\\
                \mathcal{F}_\text{eq}({z}^{*}_C,\hat{l})
    \end{array}\right]= 0.
\end{aligned}
\end{equation}
According to the differential principle of implicit function, the gradient of $\widetilde{z}$ over $\hat{l}$ can be obtained as follows:
\begin{equation}\label{implicit}
    \frac{d\widetilde{z}}{d\hat{l}}=-\mathcal{K}^{-1}_{\widetilde{z}}(\widetilde{z},\hat{l})\mathcal{K}_{\hat{l}}(\widetilde{z},\hat{l})
\end{equation}
The desired gradient $d{O}^{*}/d\hat{l}$ can be derived based on the chain rule.
The first term can be computed straightforwardly, while the second term can be derived from \eqref{implicit}.
\begin{equation}\label{gradient}
    \frac{d{O}^{*}}{d\hat{l}}=\frac{d{O}^{*}}{d{z}^{*}_C}\frac{d{z}^{*}_C}{d\hat{l}}
\end{equation}
Note that even with the help of OptNet, Steps 2 and 3 present challenges in a privacy-preserving setting, which will be discussed in detail in Section~\ref{methodology}.

\section{Methodology}
This section addresses the two questions proposed in Section~\ref{PS}. Specifically, we propose a privacy-preserving DFL framework, along with a privacy-preserving LPR algorithm for adaptive DFL to address load heterogeneity.

\label{Methodology}
\subsection{Privacy-preserving DFL}
\label{Methodology1}

\begin{figure}
    \centering
    \includegraphics[width=0.95\linewidth]{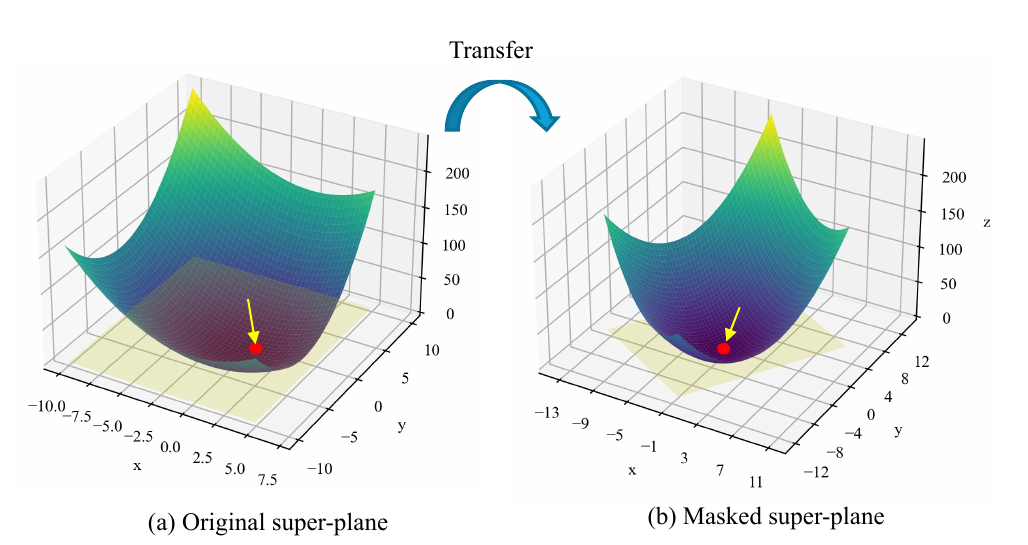}
    \caption{The basic idea of IM. The super-plane illustrates the mapping from decision variables ($x$, $y$) to the objective ($z$) for both the original and masked optimization 
    problems, where masking is achieved by scaling and rotating the variable}
    \label{illustration_IM}
\end{figure}

Existing privacy-preserving encryption methods, such as differential privacy, are mainly designed to safeguard data during querying and analytical processes \cite{wang2023data}. However, when applied to DFL tasks, these conventional techniques often struggle to guarantee both the optimality and feasibility of optimization problems under strict privacy constraints. Although certain distributed optimization techniques, such as the alternating direction method of multipliers \cite{liu2017distributed}, have been proposed to enhance privacy in decision-making, significant challenges remain. Each distributed optimization requires multiple rounds of iteration for convergence, and gradient computation must backpropagate through the entire iterative history, resulting in potential gradient vanishing. Additionally, the concurrent execution of model updates and optimization iterations creates pathological coupling, undermining theoretical convergence guarantees. Finally, sector-independent DFL implementations that prioritize local objectives often yield equilibria that fall short of system-wide optimality, while a lack of coordination further impairs convergence efficiency.

One potential solution is to convert the original optimization problem into a masked optimization problem using IM, where the parameters of these two optimization problems are different, but their optimal solution and the objective adhere to a specific relationship \cite{xin2017information, han2024privacy}. 
As shown in Fig. \ref{illustration_IM}, the red point indicates the optimal solution, the yellow arrows show the gradients, and the yellow area on the $x$-$y$ plane represents the feasible region. The figure demonstrates that, if the transformation parameters (e.g., rotation angle) are known, the original solution and gradient can be recovered from the masked problem.

\begin{figure}[t]
\centering
\includegraphics[width=0.48\textwidth]{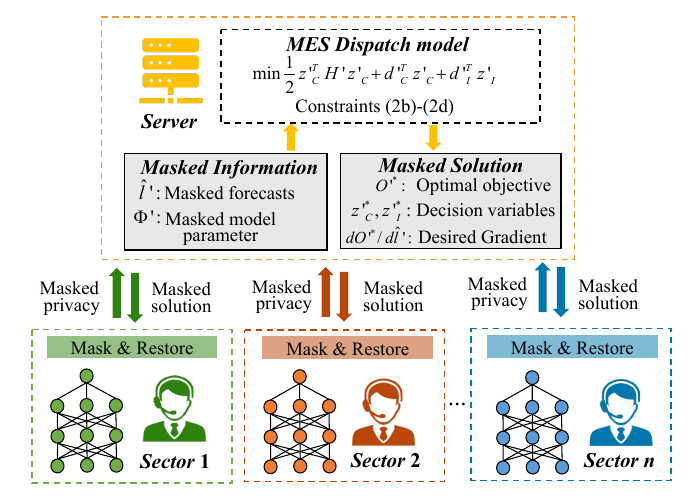}
\caption{Illustration of IM-based DFL framework}
\label{methodology}
\end{figure}

Building on this, we propose a privacy-preserving DFL framework based on IM. As illustrated in Fig.~\ref{methodology}, forward propagation enables MES dispatch without direct access to private data, while backward propagation recovers the original gradient for model updates. Additionally, a safety protocol is incorporated to prevent collusion and unauthorized access.
It is important to note that, within our proposed framework, a central MES operator is not required to schedule individual equipment. The cloud server is responsible solely for computational tasks and does not possess the authority to control specific units. This decentralized architecture is broadly applicable and has been adopted in related studies, such as \cite{deng2025privacy} and \cite{lin2025privacy}.

\subsubsection{Decision-making Module Forward Propagation} 
Building upon the previous discussion, this section elaborates on how IM safeguards privacy in Step~2 of Fig.~\ref{framework}. Specifically, IM is designed to prevent the disclosure of original load data $\hat{l}$ and sensitive parameters $\Phi = \{M, N, M_{\text{EB}}, N_{\text{EB}}, H, b, d_C, d_I\}$ throughout the solution process of the decision-making problem~\eqref{original_prob}.
The general concept of IM involves the linear transfer of the original $\Phi$ and $\hat{l}$ with the masking matrices set $\Omega=\{Q, W, G, T, T_{\text{EB}},r,u\}$ (the definition will be given as follows). 
The masked problem is represented as follows \cite{chen2023privacy}:
\begin{subequations}\label{masked_problem}
\begin{alignat}{4}
\min_{z'_C,z'_I}\quad & \frac{1}{2}{z'}_C^{\text{T}}H'{z'}_C+{d'}_{C}^{\text{T}} z'_C+{d'}_{I}^{\text{T}}z'_{I}\label{mask_eq1}\\ 
  \text{s.t.} \quad & I'z'_C\geq -I'r \label{mask_eq2}\\
      & M'z'_C+N'z_I'=b' \label{mask_eq3}\\ 
&M'_{\text{EB}}z'_C+N'_{\text{EB}}z'_I=\hat{l'} \label{mask_eq4}
\end{alignat}
\label{straincomponent}
\end{subequations}
where the masked parameters are defined as:
\begin{equation}\label{mask_matrix}
\begin{aligned}
&H'=Q^\text{T}HQ,&&d'_I=Q^\text{T}d_I+Q^\text{T}HQr\\
&d'_C=G^\text{T}W^\text{T}d,&&I'=IQ\\
&M'=TMQ, &&M'_{\text{EB}}=T_{\text{EB}}M_{\text{EB}}Q\\
&N'=TNWG, &&N'_{\text{EB}}=T_{\text{EB}}N_{\text{EB}}WG\\
&b'=Tb-M^{'}r-N^{'}u,
&&\hat{l'}=T_{\text{EB}}\hat{l}-M'_{\text{EB}}r-N'_{\text{EB}}u
\end{aligned}
\end{equation}
where $Q$, $T$, and $T_{\text{EB}}$ are non-singular matrices, $r$ is random vectors, $W$ is a monomial matrix (which has only one non-zero element in each row or column and the value of nonzero elements in $W$ is 1), $G$ is a diagonal matrix and its diagonal elements are either $1$ or $-1$,
$u$ is a vector that related to $G$, which is defined in \eqref{def_u}. Note that all the masking matrices $\Omega$ are stored locally on the side of the sectors, and the server has no access to them.
\begin{equation}  \label{def_u} 
u_i=\left\{     \begin{aligned}     1 & , \quad G_{ii}=-1\\     0 & ,\quad   G_{ii}=1   
\end{aligned}     \right. 
\end{equation}
where $u_i$ is the $i^{\text{th}}$ elements of $u$, $G_{ii}$ is the $i^{\text{th}}$ diagonal elements of $G$.
Assume the optimal continuous variables and integer variable of the original problem and masked problem are $z^{*}_C$, $z^{*}_I$, ${z'}^{*}_C$, ${z'}^{*}_I$, respectively. It has been proven that they adhere to the following relationship \cite{tian2021privacy}.
\begin{equation}\label{solution_relationship}
    z^{*}_C=Q({z'}_C^{*}+r)\quad z^{*}_I=W(G{z'}_{I}^{*}+u)
\end{equation}
By implementing this approach, sectors can restore the solution of the original problem using $\Omega$, while the server is restricted to accessing only the masked parameters. This ensures that the privacy-preserving Step 2 can be accomplished.

\subsubsection{Decision-making Module Backward Propagation}
While Section \ref{preliminaries} introduced the fundamental concepts of OptNet, two critical challenges remain for backpropagation through IM-based decision-making modules: 
\begin{enumerate}
    \item \textit{Challenge 1}: Handling of integer variables inherent in decision-making formulations
    \item \textit{Challenge 2}: Gradient recovery from masked problems to original solution spaces
\end{enumerate}
As for challenge 1, decision-making in MES inevitably involves the introduction of integer variables to describe unit commitment or to approximate nonlinear constraints. The presence of integer variables renders the optimization problem non-differentiable~\cite{wilder2019melding}.
Several attempts have been made to address this issue by identifying the optimal relaxed problem and constructing the OptNet for the relaxed formulation. For instance, \cite{ferber2020mipaal} employs cutting plane methods to search for the optimal problem, while \cite{zhou2024load} leverages the branch and bound method for this purpose.
All these integer-solving methods for DFL problems can also be directly applied to the masked problem on the server side within our framework. This is because the process of finding the optimal subproblem and the gradient recovery process are decoupled. In this work, we adopt the branch and bound method for applying OptNet to MIQP/LP problems, as it efficiently explores the solution space and guarantees global optimality~\cite{dan2020scenario,lawler1966branch}.
After obtaining the optimal subproblem with branch and bound, the relaxed version of~\eqref{masked_problem} is formulated as follows, where ${z'}_I^{*}$ denotes the integer variables of the optimal subproblem as \eqref{relaxed_prob}. For the relaxed problem, we can develop an OptNet and calculate the desired gradient based on \eqref{gradient}.
\begin{subequations}\label{relaxed_prob}
\begin{alignat}{4}
\min_{z'_C} \quad & \frac{1}{2}{z'}_C^{\text{T}}Hz_C^{'}+d_C^\text{T}{z'}_C\\ 
  \text{s.t.} \quad & I'{z'}_C\geq -{I'}r\\
      & Mz'_C=b'-N{z'}_I^{*}\\ 
&M_{\text{EB}}z'_C=\hat{l'}-N_{\text{EB}}{z'}_I^{*}
\end{alignat}
\label{straincomponent}
\end{subequations}

Regarding challenge 2, this is unique to IM-based DFL approaches. As proof in Appendix \ref{appendix}. The relationship between the gradient of the optimal dispatch cost (${O'}^{*}$ and ${O}^{*}$) over forecasts ($\hat{l'}$ and $\hat{l}$) of the masked problem and the original problem is related to $Q$ and $T$:
\begin{equation}\label{gradient_relationship1}
 \frac{\partial z_C^{*}}{\partial \hat{l}}=Q\frac{\partial {z'}_C^{*}}{\partial \hat{l'}}T_{\text{EB}}\quad
    \frac{\partial O^{*}}{\partial \hat{l}}=\frac{\partial {O'}^{*}}{\partial \hat{l'}}T_{\text{EB}}
\end{equation}
Hence, backward propagation of the decision-making module can be achieved. The server can calculate the masked desired gradient ($d{O'}^{*}/d \hat{l'}$) and transmit it to the sectors. The sectors can then restore the original desired gradient ($d{O}^{*}/d \hat{l}$) using the masking matrices $Q$ and $T_{\text{EB}}$ on their local side. 

\subsection{Safety Protocol for Privacy-preserving DFL}
The IM mechanism enables us to achieve privacy-preserving forward and backward propagation. However, sharing the masking matrices $\Omega$ in \eqref{mask_matrix} across all sectors presents potential security risks:

\textit{Risk 1:} If a malicious sector intercepts the masked parameters shared by other sectors, it can easily obtain the raw private information of other sectors with $\Omega$.

\textit{Risk 2:} If a malicious sector collaborates with the server to share the $\Omega$, the operator could gain access to the original load data $\hat{l}$ and model parameter $\Phi$ of all sectors.

\textit{Risk 3:} In DFL, even if the malicious sector only shares $T_{\text{EB}}$ with the operator, privacy related to the original desired gradient $\partial O^{*}/\partial \hat{l}$ may still be compromised. 

This work proposed a safety protocol to mitigate these risks for IM-based DFL.
Firstly, matrix blocking is introduced, where each sector $s_i$ independently designs its masking matrix $\Omega_i$. Then, HE is employed to address additional requirements specific to DFL, e.g., ensuring that $T_{\text{EB}}$ remains private, in order to protect gradient information.

\subsubsection{Matrix blocking}
Assume decision variables related to sector $s_i$ are $z^{\text{T}}_{C,i}$ and $z^{\text{T}}_{I,i}$. The decision variables $z^{\text{T}}_{C}$ and $z^{\text{T}}_{I}$ in \eqref{masked_problem} can be rewrite as:
\begin{equation}
 z^{\text{T}}_C=[z^{\text{T}}_{C,1},z^{\text{T}}_{C,2},\cdots,z^{\text{T}}_{C,n}]\quad z^{\text{T}}_I=[z^{\text{T}}_{I,1},z^{\text{T}}_{I,2},\cdots,z^{\text{T}}_{I,n}]
\end{equation}
The parameter in $\Phi$ can also be rewritten in the form of block matrices. 
\begin{equation}
\begin{aligned}
&M=\text{diag}\{M_1,M_2,\cdots,M_n\}\quad I=\text{diag}\{I_1,I_2,\cdots,I_n\}\\
&N=\text{diag}\{N_1,N_2,\cdots,N_n\}\quad H=\text{diag}\{H_1,H_2,\cdots,H_n\}\\
&M_{\text{EB}}=[M_{\text{EB},1},M_{\text{EB},2},\cdots,M_{\text{EB},n}]
\quad b^{\text{T}}=[b^{\text{T}}_{1},b^{\text{T}}_{2},\cdots,b^{\text{T}}_{n}]\\
&N_{\text{EB}}=[N_{\text{EB},1},N_{\text{EB},2},\cdots,N_{n,\text{EB},n}]
\quad d^{\text{T}}=[d^{\text{T}}_{1},d^{\text{T}}_{2},\cdots,d^{\text{T}}_{n}]\\
\end{aligned}
\end{equation}
As for risk 1 and risk 2 caused by sharing masking matrices, $\Omega$ can also be designed in the form of block matrices. This means each client $s_i$ can design their masking matrices $\Omega_i=\{Q_i, G_i,W_i,T_i,T_{\text{EB}},r_i,u_i\}$ independently.
\begin{equation}
\begin{aligned}
&Q=\text{diag}\{Q_1,Q_2,\cdots,Q_n\}
&&G=\text{diag}\{G_1,G_2,\cdots,G_n\}\\
&W=\text{diag}\{W_1,W_2,\cdots,W_n\}
&&T=\text{diag}\{T_1,T_2,\cdots,T_n\}\\
&T_{\text{EB}}=[T_{\text{EB},1},T_{\text{EB},2},\cdots,T_{\text{EB},n}]
 &&  u^{\text{T}}=[u^{\text{T}}_{1},u^{\text{T}}_{2},\cdots,u^{\text{T}}_{n}]\\
&r^{\text{T}}=[r^{\text{T}}_{1},r^{\text{T}}_{2},\cdots,r^{\text{T}}_{n}]
\end{aligned}
\end{equation}
The masked parameters $M',N',H',b',d'$ in \eqref{masked_problem} can be obtained with those parameters related to sector $s_i$ can be calculated as follows:
\begin{equation}
\begin{aligned}
&d'_{I,i}=G_i^{\text{T}}W_i^{\text{T}}d_{I,i}\quad M'_i=T_iM_iQ_i \quad 
N'_i=T_iN_iW_iG_i\\
&H'_i=Q_i^{\text{T}}H_iQ_i \quad d'_{C,i}=Q_i^{\text{T}}d_{C,i}+Q_i^{\text{T}}H_iQ_ir_i
\end{aligned}
\end{equation}
As for the construction of masked $M'_{\text{EB}}, N'_{\text{EB}}, \hat{l'}$, we will use the construction of masked
$M'_{\text{EB}}$ as an example, as the construction of $N'_{\text{EB}},\hat{l'}$ are similar.
\begin{equation}
\begin{aligned}
&M'_{\text{EB}}=T_{\text{EB}}[M_{\text{EB},1}Q_1,M_{\text{EB},2}Q_2,\cdots,M_{\text{EB},n}Q_n]\\
&=T_{\text{EB}}\left[
\begin{array}{cccc}
   (M_{\text{EB},1}Q_1)_1&(M_{\text{EB},2}Q_2)_1&\cdots&(M_{\text{EB},n}Q_n)_1\\
   (M_{\text{EB},1}Q_1)_2&(M_{\text{EB},2}Q_2)_2&\cdots&(M_{\text{EB},n}Q_n)_2\\
   \vdots&\vdots&\ddots&\vdots\\
  (M_{\text{EB},1}Q_1)_n&(M_{\text{EB},2}Q_2)_n&\cdots&(M_{\text{EB},n}Q_n)_n
\end{array}
\right]
\end{aligned}
\end{equation}
The masked parameters related to sector $k$ can be created as follows:
\begin{equation}
\begin{aligned}
M'_{\text{EB},k}=\sum_{i=1}^{n}T_{\text{EB},i}(M_{\text{EB},k}Q_k)_i
\end{aligned}
\end{equation}
As for Risk 3, the server will send the masked desired gradient $\partial {O'}^{*}/{\partial \hat{l'}}$ to each sector, and these sectors can restore the original gradient using $T_{\text{EB},i} \in \Omega_i$. 
\begin{equation}
\begin{aligned}\label{gradient_relationship_extend}
&[\frac{\partial O^{*}}{\partial \hat{l}_1},\frac{\partial O^{*}}{\partial \hat{l}_2},\cdots,\frac{\partial O^{*}}{\partial \hat{l}_n}]\\
&=[\frac{\partial {O'}^{*}}{\partial \hat{l'}}T_{\text{EB},1},\frac{\partial {O'}^{*}}{\partial \hat{l'}}T_{\text{EB},2},\cdots,\frac{\partial {O'}^{*}}{\partial \hat{l'}}T_{\text{EB},n}]
\end{aligned}
\end{equation}

\textit{HE Protocol:}
The construction of $M'_{\text{EB},k}$ requires sector $s_k$ to send different parts of $M_k Q_k$ to all other sectors $s_i$ $(i \neq k)$. Although the original information in $M_k$ is masked by $Q_k$, there still exist risks of collusion and privacy leakage.

1) If the sector $s_k$ directly send $(M_kQ_k)_i$ to sector $s_i$, although the information in $M_k$ is masked by $Q_k$, there may still be a collusion issue. Specifically, the server will obtain $T_{\text{EB},i}(M_kQ_k)_i$ directly. In this situation, if the sector $s_k$ sends $(M_kQ_k)_i$ to the server, it can know $T_{\text{EB},i}$ and restore the original gradient of sector $s_i$. 

2) When calculating the masked $N'_{\text{EB}}$, the sectors $s_k$  should share $(N_{\text{EB},k}W_kG_k)_i$ with all other sectors $s_i,i\neq k$. The elements in $W$ and $G$ are 0, 1 and -1, which do not effectively mask the privacy of the original data, potentially resulting in privacy issues.

\begin{figure}[t]
\centering
\includegraphics[width=0.45\textwidth]{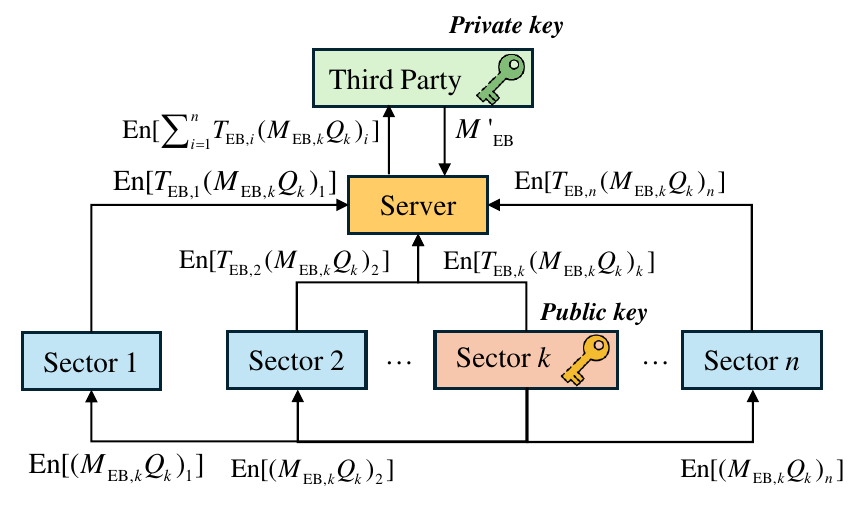}
\caption{Proposed safety protocol for privacy-preserving DFL} 
\label{protocol}
\end{figure}

Hence, HE is introduced as shown in Algorithm \ref{encryoted_alg}. The illustration of the proposed safety encryption protocol based on HE is shown in Fig. \ref{protocol}. 
We employ additively HE, which satisfies the property that:
\begin{equation}
    \textbf{En}[\sum_{i}^n T_i(M_{\text{EB},i}Q_k)_i]=\Pi_{i=1}^{n} \textbf{En}[T_i(M_{\text{EB},i}Q_k)_i]\label{he_eq}
\end{equation}
For each sector, a trusted third party generates and distributes encryption keys. Each sector encrypts its masked parameters and securely exchanges the encrypted data with other sectors, who further process and upload the results to the server. The server aggregates all encrypted results, which are then decrypted by the third party to obtain the final masked parameters.


{\setlength{\baselineskip}{0.95\baselineskip}
\begin{algorithm}[t]
\caption{Enhanced IM with the safety protocol}
\label{encryoted_alg}
\SetKwInOut{KIN}{Input}
\SetKwInOut{KOUT}{Output}
\KIN{$M_{\text{EB},i}$, $Q_i$, $T_{\text{EB},i}$ of sectors $s_i$, $i\in [1,n]$}
\KOUT{Masked parameters $M'_\text{EB}$}
  \For{$k \in [0,n]$}{
    A trustworthy third party creates a public key and a private key\;
    third-party sends public key to sector $s_k$\;
    \For{$i\in[1,n],i\neq k$}{
    sector $s_k$ encrypted $(M_{\text{EB},k}Q_k)_i$\;
    sector $s_k$ send $\textbf{En}[(M_{\text{EB},k}Q_k)_i]$ to sector $s_i$\;
    sector $s_i$ calculate $\textbf{En}[T_i(M_{\text{EB},k}Q_k)_i]$ and
    upload to operator\;
}
sector $s_k$ send $\textbf{En}[(M_{\text{EB},k}Q_k)_k]$ to server\;
server calculate $\textbf{En}[\sum_{i}^n T_i(M_{\text{EB},i}Q_k)_i]$ with \eqref{he_eq}\;
third-party decrypt $\textbf{En}[\sum_{i}^n T_i(M_{\text{EB},i}Q_k)_i]$ with private key and send back to server\;
server recieve $M'_{\text{EB},k}$\;
}
\textbf{Return}  $M'_\text{EB}=[M'_{\text{EB},1},M'_{\text{EB},2},\cdots,M'_{\text{EB},n}]$ 
\end{algorithm}

\subsection{Privacy-preserving LPR for Adaptive DFL}
Multi-energy loads exhibit strong correlations and seasonality, with varying load patterns across different seasons. Load patterns refer to the representative and regular energy consumption way over a specific period \cite{cheng2021probabilistic}. For two profiles from different load patterns, the operating status of devices in MES, such as heat and cooling storage tanks, may vary between winter and summer. 
The integer solutions $z_I^*$ in \eqref{relaxed_prob} for these two profiles are likely to be different, leading to varying impacts on DFL models.
Hence, it is essential to identify different load patterns within MES and develop customized DFL models for each pattern. Various clustering methods have been proposed for LPR, among which K-means has been widely applied in MES due to its simplicity and computational efficiency \cite{zhou2023can}. However, existing studies have seldom investigated the potential privacy risks associated with centralized K-means clustering, where sensitive load profiles must be aggregated from multiple sectors \cite{wu2024multi}. To address this gap, we propose a privacy-preserving LPR algorithm for MES, leveraging the property that orthogonal transformations in Euclidean space preserve distances.

Assume the set of daily load profiles of sectors $s_i$ is $P_i=\{l_i^{\alpha}|\alpha \in [1,|P_i|]\}$ where $|P_i|$ is the number of profiles. Firstly, load profiles are normalized due to potential significant differences, denoted as ($\overline{P}_i$ and $\overline{l}^{\alpha}_{i,t}$). 
Then, to protect the privacy of daily load profiles, each sector $s_i$ for $i\in[0,n]$ creates an orthogonal masking matrix $V_i\in \mathbb{R}^{\tau \times \tau}$ and random vectors $f_i\in \mathbb{R}^{\tau}$ independently. These matrices $V_i$ and vectors $b_i$ are utilized to mask the original normalized profiles as follows:
\begin{equation}
    \overline{{l'}}_i^{\alpha}=V_i\overline{l}_i^{\alpha}-f_i\quad \alpha \in [1,|P_i|]
\end{equation}
where $\overline{{l'}}_i^{\alpha}$ and $\overline{l_i}^{\alpha}$ are masked and original normalized profiles.
Appendix \ref{appendixb} proof that such an IM mechanism will not affect the distance between normalized profiles, thereby ensuring that the clustering result remains unchanged.
The details of the IM-based K-means are listed in Algorithm \ref{IM-kmeans}.
After the load patterns are recognized without privacy leakage, an adaptive DFL approach is developed for each load pattern to address the heterogeneity between different patterns as discussed in Subsection \ref{Methodology1}.
\setlength{\baselineskip}{0.92\baselineskip}
\begin{algorithm}[t]
\caption{IM-based K-means}\label{IM-kmeans}
\SetKwInOut{KIN}{Input}
\SetKwInOut{KOUT}{Output}
\KIN{Daily load profiles $P_i$ for $i\in [1,n]$, Minimum and maximum cluster number $CN_{\text{min}}$ and $CN_{\text{max}}$}
\KOUT{Cluster center profiles $\overline{P}_{c}$}
\SetKwProg{Fn}{IM procedure}{:}{}
\Fn{}{
  \For{$i \in [0,n]$}{
    Sector $s_i$ creates orthogonal masking matrix $V_i$ and random vectors $f_i$ independently\;
    Sector $s_i$ normalize $P_i$ and obtain $\overline{P}_i$\;
    Sector $s_i$ generate masked normalized daily load profiles $\overline{P}'_i$ with $T_i,b_i$\;
    Sector $s_i$ uploads $\overline{P'}_i$ to operator\;
  }
}
\SetKwProg{Fn}{Clustering procedure}{:}{}
\Fn{}{ 
  Server collects $\overline{P}'_i$, $i\in [1,n]$ and concatenates them into a unified set $\overline{P}'$\;
  \For{$j \in [CN_{\min},CN_{\max}]$}{
    Conduct K-means algorithms to $\overline{P}'$\;
    Record cluster center profiles $\overline{P}_{c,j}$\;
  }
  Determine the optimal cluster number $j^{*}$ with elbow method and $\overline{P}_{c}=\overline{P}_{c,{j^{*}}}$\;
}
\textbf{Return $\overline{P}_{c}$}
\end{algorithm}

\section{Case Studies}
\label{CS}
\subsection{Experiment Settings}
\subsubsection{Dataset}
We selected an open dataset named Building Data Genome Project 2 for case studies \cite{miller2020building}. This dataset contains hourly data from 2016 to 2017 collected from 19 stations across North America and Europe. In particular, we utilized data from the Bull Station, located at the University of Texas, Austin. The total electricity, cooling, and heat loads were derived by aggregating loads from different buildings. The data collected in 2016 is used as the training dataset, and 10\% of it is randomly selected as the validation dataset. The data collected in 2017 is utilized as the test dataset. In our experiments, artificial neural networks are adopted as the forecasting model, and MAPE and root mean square error (RMSE) are selected to evaluate the forecasting accuracy.

\subsubsection{MES Setting}
The studied MES has three energy sectors, i.e., electricity, heat, and cooling. The combined device in the studied MES is a combined cooling, heat, and power (CCHP) equipment. The turbine in the CCHP is a backpressure turbine, so the electricity and heat generated by the CCHP are in a certain proportion. There are also gas boilers (GB), electric boilers (EB), electric refrigerators (ER), and storage units for the three sectors.
MES dispatch includes day-ahead and intra-day stages. In the day-ahead stage, the operator schedules devices based on sector forecasts. During intra-day dispatch, actual loads may deviate from forecasts, and the operator maintains energy balance by temporarily purchasing electricity from the grid, adjusting generation within backup capacity, or utilizing storage unit charging and discharging.
The details of constraints and objective functions are given in Appendix \ref{appendixc}

In our experiments, artificial neural networks are adopted as the forecasting model. The model incorporates three types of input features: historical features (loads of corresponding sectors on the previous day), predicted weather features (temperature, humidity), and calendar features (week, month). The output of the forecasting model is the load profiles for the following day. The detailed parameters of the case study are given in Table \ref{table0}.

\begin{table}[t]
\centering
\caption{Parameters of case study}
\begin{tabular}{cc}
\hline
Parameter                        & Value     \\ \hline
Maximum of forecasting model training & 1000 \\
 Early stopping patience & 10       \\
Batch size                        & 32 \\
learning rate of forecasting                     & 1e-03 \\ 
Number of hidden layers             & 3   \\  
Number of neuron nodes in each layer             & 64       \\
Epoch of DFL training & 5\\
Learning rate of DFL & 2e-5\\
Optimizer                          & Adam\\  Normalization                   & Max-min \\ \hline
\end{tabular}\label{table0}
\end{table}

\subsection{Results}
\begin{figure}[t]
\centering
\includegraphics[width=0.45\textwidth]{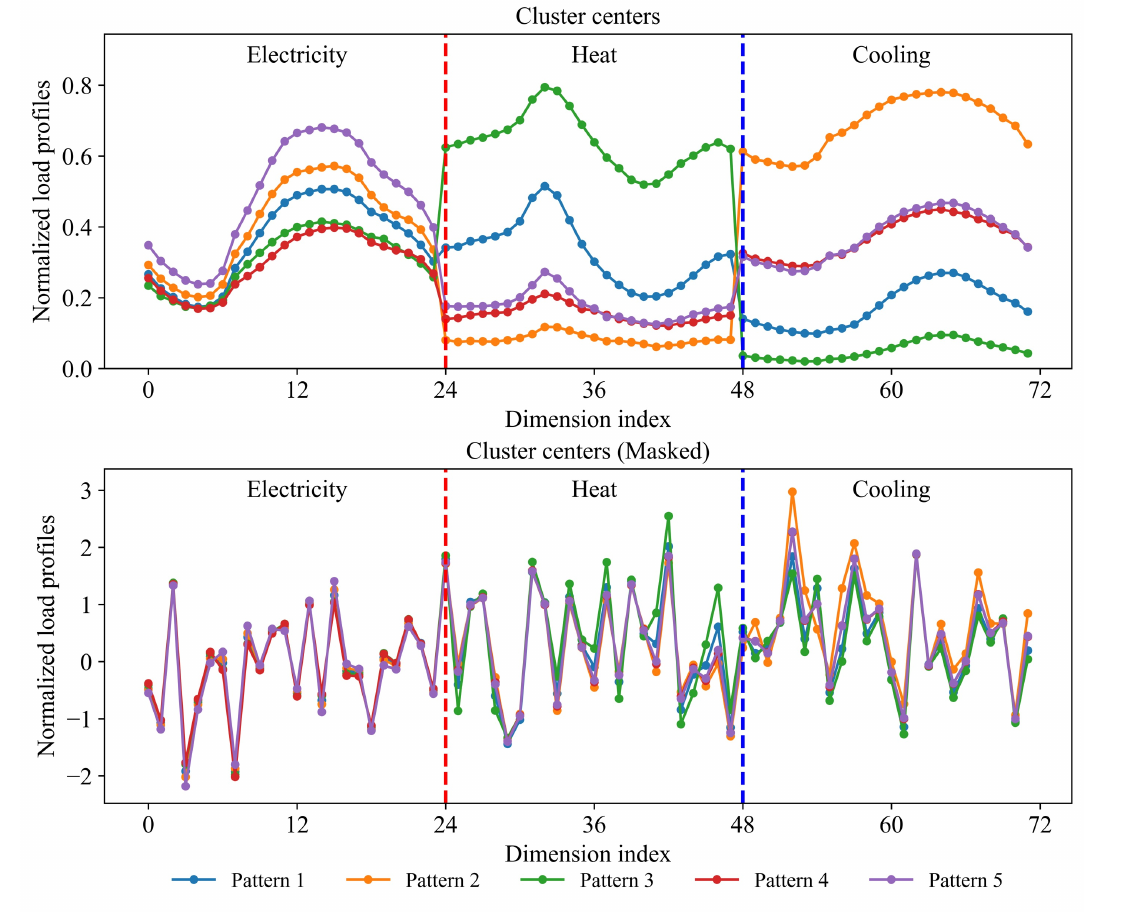}
\caption{Original and masked cluster centers } 
\label{cluster_center}
\end{figure}
In this part, we will evaluate our proposed adaptive DFL (ADFL) compared to the traditional model trained with MAE (Benchmark), standard DFL (SDFL), which trains a single model for all patterns.
Before training the forecasting with a decision-making module, a pre-training step is required to train basic models, and the pre-training procedure will be stopped by early stopping.
In this work, the optimal cluster number is 5, which is determined by the elbow method.
The original and masked cluster centers are depicted in Fig. \ref{cluster_center}. The $x$-axis represents the dimension index of the electricity, heat, and cooling load profiles, with 0-24, 24-48, and 48-72 corresponding to the normalized electricity, heat, and cooling loads, respectively. In the traditional K-means, we can analyze the user behavior associated with each pattern. For example, in summer, there may be situations where the cooling load is high and the heat load is low (patterns 3 and 4). However, this information cannot be obtained in the IM-based K-means approach because all profiles are masked, and the cluster centers shown to the operator appear jumbled up.

We have conducted five repeat experiments with different initial seeds. The average forecasting accuracy of sectors in MES and the average daily dispatch costs of different methods are listed in Table \ref{table1}. 
The DFL approach shows a decrease in forecasting accuracy compared to traditional training methods for the three sectors in MES. Since forecasting is used for decision-making, this study will prioritize analyzing the dispatch costs.
The result shows that the average daily dispatch of DFL methods (SDFL, ADFL) is lower than the benchmark. This phenomenon emphasizes the discrepancy between minimizing forecasting and minimizing operation costs. Although the relative percentage decrease may not be very high, this is because generating energy that matches the load requires expensive costs. Therefore, even a small ratio in cost savings can lead to significant cost reductions over the entire year.
The results suggest that our proposed ADFL methods can further reduce daily dispatch costs across various load patterns throughout the entire year of 2017 compared to the DFL method.
Compared to the DFL method, which saves 240.097 kCNY every year over the benchmark, our approach can achieve savings of 307.075 kCNY, with a cost-saving increase percentage of 21.81\%. It shows that clustering different load patterns effectively mitigates the impact of heterogeneity on the cost reduction performance of DFL.

\begin{table*}[htb]
\caption{Forecasting accuracy and dispatch cost of different models}
\centering
\label{table1}
\begin{tabular}{ccccccccc}
\hline
                           &           & \multicolumn{3}{c}{MAPE}                                                                  & \multicolumn{3}{c}{RMSE}                                                                  & \multicolumn{1}{c}{\multirow{2}{*}{Cost (kCNY)}} \\ \cline{3-8}
                           &           & \multicolumn{1}{c}{Electricity} & \multicolumn{1}{c}{Heat} & \multicolumn{1}{c}{Cooling} & \multicolumn{1}{c}{Electricity} & \multicolumn{1}{c}{Heat} & \multicolumn{1}{c}{Cooling} & \multicolumn{1}{c}{}                             \\ \hline
\multirow{3}{*}{Pattern 1} & Benchmark & 0.0270                          & 0.0941                    & 0.1763                      & 84.6100                         & 206.2164                  & 564.2251                    & 52.7068                                          \\
                           & SDFL       & 0.0297                          & 0.1210                    & 0.1863                      & 91.6319                         & 237.1202                  & 630.8655                    & 52.1939                                          \\
                           & ADFL     & 0.0293                          & 0.1319                    & 0.1822                      & 90.3541                         & 251.0686                  & 609.7509                    & \textbf{52.1225}                                          \\ \hline
\multirow{3}{*}{Pattern 2} & Benchmark & 0.0322                          & 0.0640                    & 0.0611                      & 114.1330                        & 67.0294                   & 523.7657                    & 69.9156                                          \\
                           & SDFL       & 0.0374                          & 0.0881                    & 0.1014                      & 124.5984                        & 90.9255                   & 818.6653                    & 69.0908                                          \\
                           & ADFL     & 0.0400                          & 0.1058                    & 0.1300                      & 130.0355                        & 105.7646                  & 1021.3250                   & \textbf{68.8090}                                          \\ \hline
\multirow{3}{*}{Pattern 3} & Benchmark & 0.0301                          & 0.0986                    & 0.1669                      & 93.2497                         & 308.4534                  & 309.4917                    & 50.8096                                          \\
                           & SDFL       & 0.0327                          & 0.1298                    & 0.1635                      & 100.0067                        & 374.0575                  & 328.0463                    & 50.7766                                          \\
                           & ADFL     & 0.0302                          & 0.1014                    & 0.1674                      & 93.5183                         & 314.1398                  & 310.9290                    & 50.7983                                          \\ \hline
\multirow{3}{*}{Pattern 4} & Benchmark & 0.0251                          & 0.0655                    & 0.1218                      & 69.2440                         & 91.0299                   & 653.9543                    & 57.3018                                          \\
                           & SDFL       & 0.0302                          & 0.0958                    & 0.1488                      & 82.7724                         & 126.9397                  & 792.7847                    & 56.5044                                          \\
                           & ADFL     & 0.0316                          & 0.1389                    & 0.1541                      & 86.3373                         & 172.7612                  & 820.3173                    & \textbf{56.2195}                                          \\ \hline
\multirow{3}{*}{Pattern 5} & Benchmark & 0.0257                          & 0.0687                    & 0.1224                      & 92.7135                         & 99.4353                   & 686.6852                    & 62.8011                                          \\
                           & SDFL       & 0.0308                          & 0.0950                    & 0.1413                      & 106.0147                        & 125.8183                  & 798.3356                    & 62.0667                                          \\
                           & ADFL     & 0.0324                          & 0.1135                    & 0.1455                      & 110.3510                        & 147.3648                  & 821.9970                    & \textbf{61.8892}                                          \\ \hline
\multirow{3}{*}{Summary}   & Benchmark & 0.0283                          & 0.0749                    & 0.1188                      & 95.6675                         & 151.4105                  & 580.9511                    & 60.9784                                          \\
                           & SDFL       & 0.0327                          & 0.1017                    & 0.1415                      & 106.0373                        & 183.8462                  & 739.4114                    & 60.3206                                          \\
                           & ADFL     & 0.0338                          & 0.1173                    & 0.1516                      & 108.7432                        & 188.6007                  & 819.7347                    & \textbf{60.1371}                                          \\ \hline
\end{tabular}
\end{table*}

\subsection{Result Analysis}
\subsubsection{Visual Analysis of dispatch result}
To further explain how lower forecasting accuracy can still result in reduced dispatch costs, the study chose a specific day from the testing dataset. The forecasts of three methods and actual load profiles of the selected example day as shown in Fig. \ref{loads}. The heat load forecasting provided by ADFL is larger compared to the benchmark and DFL, while the cooling load forecasts are lower than the other two methods. Additionally, the electricity load forecasts from all three methods are similar and closely match the actual load. Therefore, we focus on supply and demand analysis within the heat and cooling sector. 

The outputs of different devices in MES are illustrated in Fig. \ref{output}, and TOU in this figure means Time-of-Use (TOU) price.
In our experiment setting, the electricity, heat, and cooling generated by CCHP are proportional. Therefore, in Fig. \ref{output}, we only provide the cooling output of CCHP. Since the heat loads on the selected day are significantly smaller than the cooling and electricity loads, the output of CCHP is determined by the heat loads on that day. Hence, in day-ahead dispatch, the CCHP output of ADFL will be slightly higher than that of the other two methods. As the CCHP output is relatively low, the cooling loads are primarily satisfied by the ER. The ER output of the proposed method is also relatively small due to the forecasts from ADFL compared to other methods.
In intra-day dispatch, various strategies can be used to address the under-forecasts of cooling loads. One method is to charge the battery during periods of low electricity prices and discharge it during periods of higher prices to power the ER and meet cooling demands. The second approach is to increase the output of the CCHP during times of high electricity prices to satisfy the cooling load. By increasing the CCHP output, the heat generated by the GB becomes significantly reduced during phases of high electricity prices. Furthermore, the cooling loads during the high electricity consumption period can also be transferred to periods with lower electricity prices with the cooling storage tank. Therefore, reducing the forecasts of cooling loads within the appropriate interval can provide more flexibility for intra-day dispatch, resulting in lower costs.
\subsubsection{Computation Complexity}

In terms of masked problem construction, generating the masking matrix $\Omega$ and constructing $M',N',H',b',d'$ are relatively quick, while the most time-consuming procedure is the construction of $M'_{\text{EB}}$. The computation complexity of construction of $M'_{\text{EB}}$ is $\mathcal{O}_c(nm_c)$. Our experiments show that for $n=3,m_c=564$, 
 the construction of $M'_{\text{EB}}$ only takes several seconds when done once.
In our setting, $N_{\text{EB}}$ is a zero matrix because there are no integer variables related to the energy balance constraints, so there is no need to construct $N'_{\text{EB}}$ in our setting.

Two additional points indicate our safety protocol will not significantly impact our training process. Firstly, the safety training is only applied during the model training process, and the time-consuming requirement is not high. Secondly, the construction of $M'_{\text{EB}}$ can be calculated in parallel for various sectors. Thirdly, in a batch training setting, the construction of $M'_{\text{EB}}$ only needs to be done once for batch training profiles. Hence, the time-consuming nature of the safety protocol will not impact the application potential of our methods.

\begin{figure}[tb]
\centering
\includegraphics[width=0.45\textwidth]{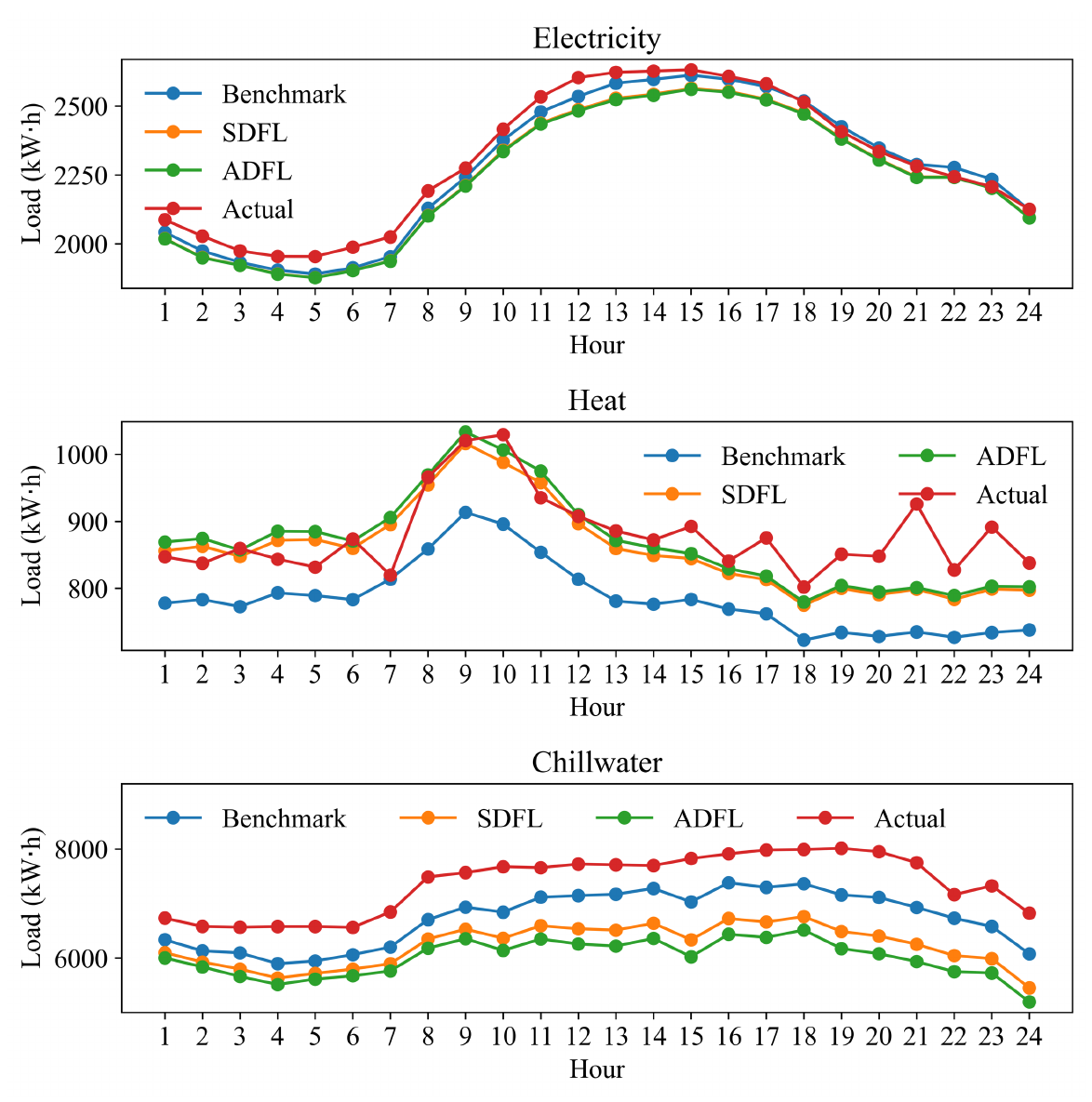}
\caption{Forecasts and actual load of a selected example day} 
\label{loads}
\end{figure}

\section{Conclusions and Future Works}
\label{conclusion}

This work proposes a privacy-preserving DFL framework for MES based on the IM mechanism. A safety protocol has been designed to prevent potential collusion between malicious clients and the server. To ensure that the DFL model can adapt to various load patterns and prevent samples from different patterns from influencing each other, this work introduces an adaptive DFL approach. The effectiveness of the proposed privacy-preserving DFL framework and load pattern recognition with IM-based K-means has been demonstrated through theoretical analysis and case studies. The experiment conducted on an open-source dataset shows that our proposed adaptive DFL effectively reduces dispatch costs. Visual analysis has been provided to offer insight into why forecasts given by proposed methods can lead to lower costs. 

\begin{figure*}[tb]
\centering
\includegraphics[width=0.95\textwidth]{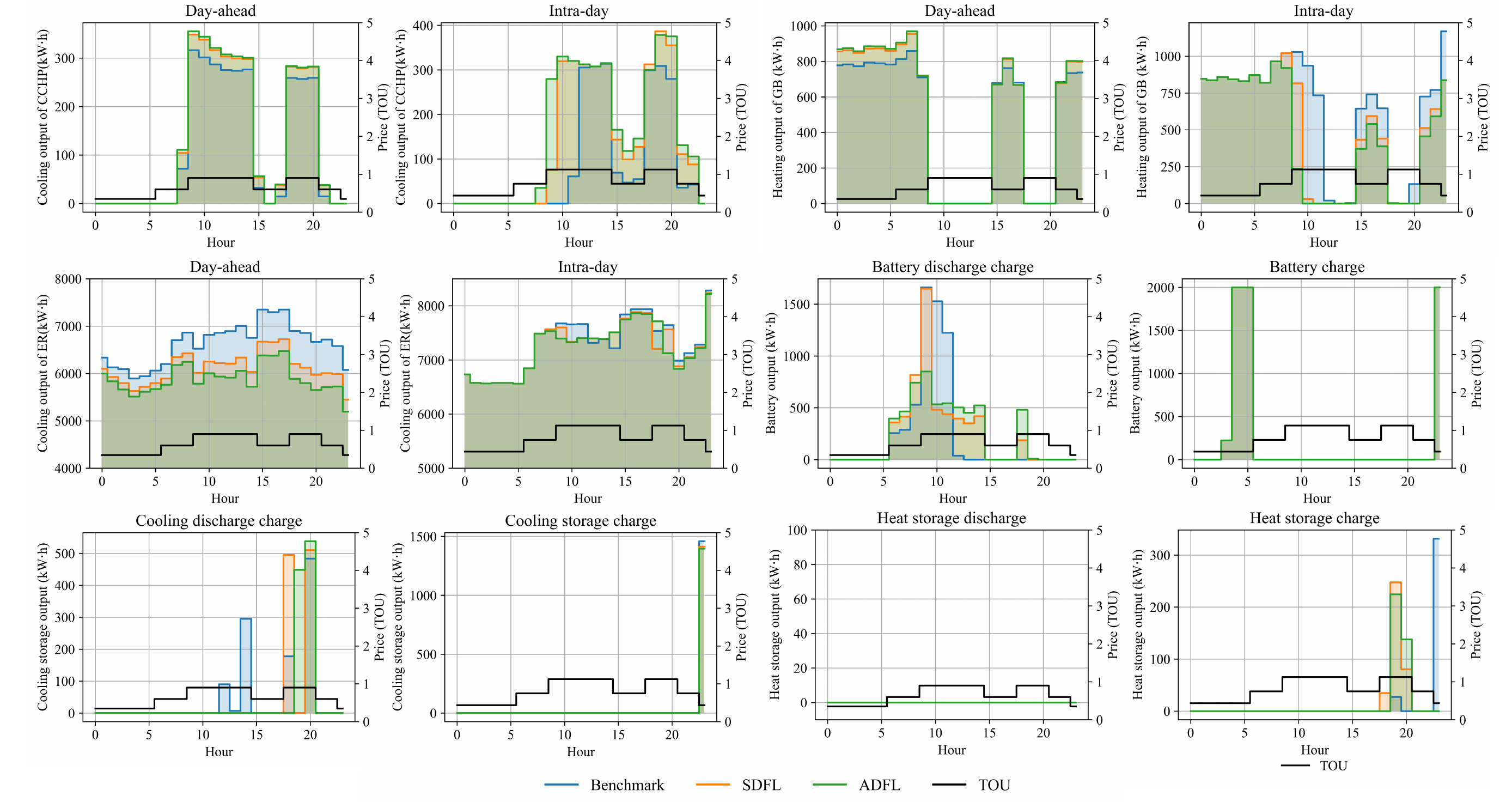}
\caption{The outputs of different devices in MES} 
\label{output}
\end{figure*}

\appendix
\label{appendix}
\subsection{IM-Based Desired Gradient Calculation}
The relationship of $\frac{\partial z_C^{*}}{\partial b}$ and $\frac{\partial {z'}_C^{*}}{\partial \hat{l'}}$ can be derived with the chain rule:
\begin{equation}
\begin{aligned}
\label{appenidx_eq1}
\frac{\partial z_C^{*}}{\partial b}= \frac{\partial z_C^{*}}{\partial {z'}_C^{*}}\frac{\partial {z'}_C^{*}}{\partial  \hat{l'}}\frac{\partial \hat{l'}}{\partial \hat{l}}&=Q\frac{\partial {z'}_C^{*}}{\partial \hat{l'}}T_{\text{EB}}
\end{aligned}
\end{equation}
$\frac{\partial {O'}^{*}}{\partial b^{'}}$ and $ \frac{\partial O^{*}}{\partial b}$ can be split into two parts based on the chain rule. The relationship between the original problem and the masked problem can be derived based on \eqref{appenidx_eq1}.
\begin{equation}
\begin{aligned}
\label{appenidx_eq2}
\frac{\partial {O'}^{*}}{\partial \hat{l'}}=
\frac{\partial {O'}^{*}}{\partial {z'}_C^{*}}\frac{\partial {z'}_C^{*}}{\partial \hat{l'}}=[\frac{1}{2}{z'}_C^{*T}({H'}^{T}+H^{'})+{d'}_C^{T}]\frac{\partial z_C^{*}}{\partial  \hat{l}}
\end{aligned}
\end{equation}
\begin{equation}
\begin{aligned}
    &\frac{\partial O^{*}}{\partial  \hat{l}}
    =\frac{\partial O^{*}}{\partial z_C^{*}}\frac{\partial z_C^{*}}{\partial  \hat{l}}=[\frac{1}{2}z_C^{*T}(H^{\text{T}}+H)+{d}_C^{\text{T}}]\frac{\partial z_C^{*}}{\partial  \hat{l}}\\
     &=[\frac{1}{2}({z'}_C^{*}+r)^{\text{T}}Q^{\text{T}}(H^{\text{T}}+H)+({d'}_C-H'r)^{\text{T}}Q^{-1}]Q\frac{\partial z_C^{*}}{\partial  \hat{l}}T_{\text{EB}}\\
     &=[\frac{1}{2}({z'}_C^{*}+r)^{\text{T}}({H'}^{T}+H')+({d'}_C-{H'}r)^{\text{T}}]\frac{\partial z_C^{*}}{\partial  \hat{l}}T_{\text{EB}}\\
     &=[\frac{1}{2}{z'}_C^{*}({H'}^{T}+H')+\frac{1}{2}r^{\text{T}}({H'}^{T}+H')+{d'}_C^{T}-(H'r)^{\text{T}}]\\
     &=[\frac{1}{2}{z'}_C^{*}({H'}^{T}+H')+{d'}_C^{T}]\frac{\partial z_C^{*}}{\partial  \hat{l}}T_{\text{EB}}=\frac{\partial {O'}^{*}}{\partial  \hat{l'}}T_{\text{EB}}
\end{aligned}
\end{equation}
\subsection{IM-based K-means}\label{appendixb}
Assume there are two samples $\alpha$ and $\beta$ their combined loads are $(l^{\alpha})^{\text{T}}=[(l_{1}^{\alpha})^{\text{T}},(l_{2}^{\alpha})^{\text{T}},\cdots,(l_{n}^{\alpha})^{\text{T}}]$,  $(l^{\beta})^{\text{T}}=[(l_{1}^{\beta})^{\text{T}},(l_{2}^{\beta})^{\text{T}},\cdots,(l_{n}^{\beta})^{\text{T}}]$. Their original squared distance is $D^2_{\alpha,\beta}=\sum_{i=1}^{n}||l_{i}^{\alpha}-l_{i}^{\beta}||^2_2$. Assume the masked concatenated loads are 
\begin{equation}
\begin{aligned}
         ({l'}^{\alpha})^{\text{T}}=[(V_1l^{\alpha}_{1}-f_1)^{\text{T}},(V_2l^{\alpha}_{2}-f_2)^{\text{T}},\cdots,(V_nl^{\alpha}_{n}-f_n)^{\text{T}}] \\({l'}^{\beta})^{\text{T}}=[(V_1l^{\beta}_{1}-f_1)^{\text{T}},(V_2l^{\beta}_{2}-f_2)^{\text{T}},\cdots,(V_nl^{\beta}_{n}-f_n)^{\text{T}}]
\end{aligned}
\end{equation}
Rewrite masking matrics $V_i$ into the formula as $V_i=[V_{i,1},V_{i,2},\cdots,V_{i,\tau}]$. The squared distance between the masked profiles $\alpha$ and $\beta$ is: 
\begin{equation}
\begin{aligned}
    D'^2_{\alpha,\beta}&=\sum_{i=1}^{n}||(V_il^{\alpha}_{i}-f_i)-(V_il^{\beta}_{i}-f_i)||^2_2\\
    &=\sum_{i=1}^{n}||\sum_{t=1}^{\tau}V_{i,t}(l^{\alpha}_{i,t}-l^{\beta}_{i,t})||^2_2\quad (V_{i,t}T^V_{j,t}=0, i\neq j)\\
    &=\sum_{i=1}^{n}\sum_{t=1}^{\tau}||V_{i,t}||^2_2(l^{\alpha}_{i,t}-l^{\beta}_{i,t})^2\\
    &=\sum_{i=1}^{n}\sum_{t=1}^{\tau}(l^{\alpha}_{i,t}-l^{\beta}_{i,t})^2=D^2_{\alpha,\beta}
\end{aligned}
\end{equation}
Hence, the distance between two samples will not be changed with the proposed IM mechanism, ensuring that the clustering results remain unchanged.
\subsection{Details of the decision-making problems in MES}\label{appendixc}
\textbf{Energy generators:} As previously mentioned, the studied MES contains various generation units. Among them, the most complex is the CCHP system, which is therefore selected as a representative example to illustrate the modeling of energy generators. The CCHP system consumes natural gas to generate electricity using backpressure gas burning turbines (TB), while the recovered heat is utilized for the production of heating and cooling through heat recovery units (HR) and absorption chillers (AR).
\begin{equation}
\begin{aligned}
    &E^{\text{ele,cchp}}_t = \gamma^{\text{tb}} q^{\text{gas}} z^{\text{cchp}}_t, \quad 
    E^{\text{ele,cchp}}_t \in [\underline{E}^{\text{ele,cchp}}, \overline{E}^{\text{ele,cchp}}] \\
    &E^{\text{heat,cchp}}_t = \gamma^{\text{rh}} \theta q^{\text{gas}}, \quad 
    E^{\text{heat,cchp}}_t \in [\underline{E}^{\text{heat,cchp}}, \overline{E}^{\text{heat,cchp}}] \\
    &E^{\text{cool,cchp}}_t = \gamma^{\text{ar}} (1-\theta) q^{\text{gas}}, \quad 
    E^{\text{cool,cchp}}_t \in [\underline{E}^{\text{cool,cchp}}, \overline{E}^{\text{cool,cchp}}]\nonumber
\end{aligned}
\end{equation}
\begin{equation}
\begin{aligned}
&-R^{\text{tb},-} \leq E^{\text{ele,cchp}}_t - E^{\text{ele,cchp}}_{t-1} \leq R^{\text{tb},+} \\
     &-R^{\text{rh},-} \leq E^{\text{heat,cchp}}_t - E^{\text{heat,cchp}}_{t-1} \leq R^{\text{rh},+} \\
     &-R^{\text{ar},-} \leq E^{\text{cool,cchp}}_t - E^{\text{cool,cchp}}_{t-1} \leq R^{\text{ar},+}
 \end{aligned}
 \end{equation}
where $\gamma^{\text{tb}}$, $\gamma^{\text{rh}}$, and $\gamma^{\text{ar}}$ are the efficiencies of TB, RH, and AR in CCHP, and $\theta$ is the fraction of recovered heat used by HR. $\underline{E}^{\text{ele,cchp}}$, $\underline{E}^{\text{heat,cchp}}$, $\underline{E}^{\text{cool,cchp}}$ and $\overline{E}^{\text{ele,cchp}}$, $\overline{E}^{\text{heat,cchp}}$, $\overline{E}^{\text{cool,cchp}}$ denote the lower and upper bounds of electricity, heat, and cooling generation, respectively. $R^{tb,-}$, $R^{rh,-}$, $R^{ac,-}$ and $R^{tb}$, $R^{rh}$, $R^{ac}$ are the downward and upward ramp limits for TB, RH, and AR.
GB, EB, and ER units are defined similarly, with $z^{\text{gb}}$, $z^{\text{eb}}$, $z^{\text{er}}$ as inputs and $E^{\text{gb}}$, $E^{\text{eb}}$, $E^{\text{er}}$ as outputs, each subject to their own bounds and ramping constraints.

\textbf{Energy Storage:}
Energy storage units for electricity, heating, and cooling are included to maintain the intra-day energy balance. For the battery:
\begin{equation}
\begin{aligned}
    &\text{SoC}_t = \text{SoC}_{t-1} 
    + \frac{\eta\, s_t^{\text{ele,char}} E_t^{\text{ele,char}}}{C^{\text{battery}}}
    - \frac{s_t^{\text{ele,dis}} E_t^{\text{ele,dis}}}{\eta\, C^{\text{battery}}} \\
    &s_t^{\text{ele,dis}} + s_t^{\text{ele,char}} = 1, \quad 
    s_t^{\text{ele,dis}} \in [0,1], s_t^{\text{ele,char}} \in [0,1] \\
    &E_t^{\text{ele,char}} \in [\underline{E}_t^{\text{ele,char}},\, \overline{E}_t^{\text{ele,char}}], \quad
    E_t^{\text{ele,dis}} \in [\underline{E}_t^{\text{ele,dis}},\, \overline{E}_t^{\text{ele,dis}}] \\
    &\text{SoC}_T = \text{SoC}_0
\end{aligned}
\end{equation}
where $\text{SoC}_t$ is the battery state of charge at time $t$, $s_t^{\text{ele,char}}$ and $s_t^{\text{ele,dis}}$ indicate charging and discharging states,
Heat and cooling storage are modeled similarly, and their outputs are denoted as $E_t^{\text{heat,char}}$, $E_t^{\text{heat,dis}}$, $E_t^{\text{cool,char}}$, $E_t^{\text{cool,dis}}$.

In the day-ahead dispatch state, the energy balance relationship constraints are provided as follows:
\begin{equation}
    \begin{aligned}
       E_t^{\text{ele,cchp}} + E_t^{\text{ele,external}} =  \hat{l}_t^{\text{ele}} + z_t^{\text{eb}} + z_t^{\text{er}} \\
        E_t^{\text{gb}} + E_t^{\text{rh}} = \hat{l}_t^{\text{heat}},E_t^{\text{ar}} + E_t^{\text{er}} =  \hat{l}_t^{\text{cool}}
    \end{aligned}
\end{equation}
The objective function of the day-ahead dispatch is formulated as:
\begin{equation}
    \begin{aligned}
        \min \sum_{t=0}^{T} \left[ h_t^{\text{ele,da}} E_t^{\text{ele,external}} + h_t^{\text{gas}} \left( z_t^{\text{gb}} + z_t^{\text{cchp}} \right) \right]
    \end{aligned}
\end{equation}
where $h_t^{\text{ele,da}}$ and $h_t^{\text{gas}}$ is price of day-ahead electricity and gas.

In the intra-day stage, there are three approaches to maintaining energy balance, as mentioned before.
\begin{equation}
    \begin{aligned}
        &\widetilde{E}_t^{\text{ele,cchp}} + E_t^{\text{ele,external}} + \widetilde{E}_t^{\text{ele,external}} + \overline{E}_t^{\text{ele,dis}} \\
        &= l_t^{\text{ele}} + \widetilde{z}_t^{\text{eb}} + \widetilde{z}_t^{\text{er}} + E_t^{\text{ele,char}} \\
        &\widetilde{E}_t^{\text{gb}} + \widetilde{E}_t^{\text{rh}} + \overline{E}_t^{\text{heat,dis}} 
        = l_t^{\text{heat}} + E_t^{\text{heat,char}} \\
        &\widetilde{E}_t^{\text{ar}} + \widetilde{E}_t^{\text{er}} + \overline{E}_t^{\text{cool,dis}}
        = l_t^{\text{cool}} + E_t^{\text{cool,char}}
    \end{aligned}
\end{equation}
where $\widetilde{E}(\cdot)$ is the adjust outout.
The objective function of the intra-day dispatch is formulated as:
\begin{equation}
    \begin{aligned}
        &\min \sum_{t=0}^{T} \Big[
             h_t^{\text{ele,rt}}\, \widetilde{E}_t^{\text{ele,external}} 
            + h_t^{\text{gas}} \left( \widetilde{z}_t^{\text{gb}} + \widetilde{z}_t^{\text{cchp}} \right) \\
            & + h_t^{\text{ele,dis}}\, E_t^{\text{ele,dis}}
              + h_t^{\text{heat,dis}}\, E_t^{\text{heat,dis}}
              + h_t^{\text{cool,dis}}\, E_t^{\text{cool,dis}} \\
            & + h_t^{\text{ele,char}}\, E_t^{\text{ele,char}}
              + h_t^{\text{heat,char}}\, E_t^{\text{heat,char}}
              + h_t^{\text{cool,char}}\, E_t^{\text{cool,char}}
        \Big]
    \end{aligned}
\end{equation}
where $h_t^{\text{ele,dis}}$,$h_t^{\text{ele,char}}$,$h_t^{\text{heat,dis}}$,$h_t^{\text{heat,char}}$,$h_t^{\text{cool,dis}}$,and $h_t^{\text{cool,char}}$ are costs related to discharging and charging of storage, $h_t^{\text{ele,rt}}$ is real-time electricity price.

\bibliographystyle{IEEEtran}
 \bibliographystyle{elsarticle-num} 

\end{document}